\documentclass[runningheads,a4paper]{llncs}

\usepackage{amssymb}
\usepackage{graphicx}
\usepackage{algorithm}
\usepackage[noend]{algpseudocode}
\usepackage{color}

\usepackage{url}
\urldef{\mailsa}\path|{eeskimez}@ur.rochester.edu,|
\urldef{\mailsb}\path|{ross.maddox, chenliang.xu, zhiyao.duan}@rochester.edu|  
\newcommand{\keywords}[1]{\par\addvspace\baselineskip
\noindent\keywordname\enspace\ignorespaces#1}

\begin{document}
\mainmatter  

\title{Generating Talking Face Landmarks from Speech}

\titlerunning{Generating Talking Face Landmarks from Speech}

%
%
\author{Sefik Emre Eskimez%
\and Ross K Maddox \and Chenliang Xu \and Zhiyao Duan \thanks{This is a pre-print of an article whose final form will appear in LVA ICA 2018. This work is supported by the University of Rochester Pilot Award Program in AR/VR and the National Science Foundation grant No. 1741471.}}
\authorrunning{Generating Talking Face Landmarks from Speech}

\institute{University of Rochester, 500 Joseph C. Wilson Blvd., Rochester, NY 14627, USA\\
\mailsa\\
\mailsb\\
}

%
%

\toctitle{Generating Talking Face Landmarks from Speech}
\maketitle

\begin{abstract}
The presence of a corresponding talking face has been shown to significantly improve speech intelligibility in noisy conditions and for hearing impaired population. In this paper, we present a system that can generate landmark points of a talking face from an acoustic speech in real time. The system uses a long short-term memory (LSTM) network and is trained on frontal videos of 27 different speakers with automatically extracted face landmarks. After training, it can produce talking face landmarks from the acoustic speech of unseen speakers and utterances. The training phase contains three key steps. We first transform landmarks of the first video frame to pin the two eye points into two predefined locations and apply the same transformation on all of the following video frames. We then remove the identity information by transforming the landmarks into a mean face shape across the entire training dataset. Finally, we train an LSTM network that takes the first- and second-order temporal differences of the log-mel spectrogram as input to predict face landmarks in each frame. We evaluate our system using the mean-squared error (MSE) loss of landmarks of lips between predicted and ground-truth landmarks as well as their first- and second-order temporal differences. We further evaluate our system by conducting subjective tests, where the subjects try to distinguish the real and fake videos of talking face landmarks. Both tests show promising results.
    
\keywords{Visual generation, face landmarks, audio-visual models, LSTM}
\end{abstract}

\section{Introduction}

	Speech is a natural way of communication, and understanding speech is essential in daily life. The auditory system, however, is not the only sensory system involved in understanding speech. The visual cues from a talker's face and articulators (lips, teeth, tongue) are also important for speech comprehension. Trained professionals are able to understand what is being said by purely looking at lip movements (lip reading) \cite{dodd1987hearing}. For ordinary people and the hearing impaired population, the presence of visual signals of speech has been shown to significantly improve speech comprehension, even if the visual signals are synthetic \cite{maddox2015auditory}. The benefits of adding the visual speech signals are more pronounced when the acoustic signal is degraded, due to background noise, communication channel distortion, and reverberation.
    
    In many scenarios such as telephony, however, speech communication is still acoustical. The absence of the visual modality can be due to the lack of cameras, the limited bandwidth of communication channels, or privacy concerns. One way to improve speech comprehension in these scenarios is to synthesize a talking face from the acoustic speech in real time at the receiver's side. A key challenge of this approach is to make sure that the generated visual signals, especially the lip movements, well coordinate with the acoustic signals, as otherwise more confusions will be introduced.

In this paper, we propose to use a long short-term memory (LSTM) network to generate landmarks of a talking face from acoustic speech. This network is trained on frontal videos of 27 different speakers of the Grid audio-visual corpus \cite{cooke2006audio}, with the face landmarks extracted using the Dlib toolkit \cite{dlib09}. The network takes the first- and second-order temporal differences of the log-mel spectra as the input, and outputs the x and y coordinates of 68 landmark points. To help the network capture the audio-visual coordination instead of the variation of face shapes across different people, we transform all training landmarks to those of a mean face across all talkers in the training set. After training, the network is able to generate face landmarks from an unseen utterance of an unseen talker. Objective evaluations of the generation quality are conducted on the LDC Audiovisual Database of Spoken American English dataset \cite{richie2009audiovisual}, which will be referred as the LDC dataset in the remaining of the paper. Subjective evaluation is also conducted to ask evaluators to distinguish speech videos with ground-truth and generated landmarks. Both the objective and subjective evaluations achieve promising results. The code and pre-trained talking face models are released to the community\footnote{\url{http://www.ece.rochester.edu/projects/air/projects/talkingface.html}}

The remaining of the paper is structured as follows: Section 2 describes the related work. Section 3 describes the data and pre-processing steps. The architecture of the network is described in Section 4. Objective and Subjective evaluations are presented in Section 5. Finally, Section 6 concludes the paper.

\section{Related Work}

Generating a talking head automatically has been a great interest in the research community. Some researchers focused on text-driven generation \cite{wang2011text,wan2013photo,fan2015photo,CASSIDY2016193}. These methods map phonemes to talking face images. Compared to text, voice signals are surface-level signals that are more difficult to parse. Besides, voices of the same text show large variations across speakers, accents, emotions, and the recording environments. On the other hand, speech signals provide richer cues for generating natural talking faces. For text, any plausible face image sequence is sufficient to establish natural communication. For speech, it must be a plausible sequence that matches the speech audio. Therefore, text-driven generation and speech-driven generation are different problems and may require different approaches.

There exist a few approaches to speech-driven talking face generation. Early work in this field mostly used Hidden Markov Models (HMM) to model the correspondence between speech and facial movements \cite{brand1999voice,choi2001hidden,cosker2003video,cosker2004speech,xie2007coupled,terissi2008audio,zhang2013new}. One of the notable early work, Voice Puppetry \cite{brand1999voice}, proposed an HMM-based talking face generation that is driven by only speech signal. In another work, Cosker et al. \cite{cosker2003video,cosker2004speech} proposed a hierarchical model that animates sub-areas of the face independently from speech and merges them into a full talking face video. Xie et al. \cite{xie2007coupled} proposed coupled HMMs (cHMMs) to model audio-visual asynchrony. Choi et al. \cite{choi2001hidden} and Terissi et. al \cite{terissi2008audio} used HMM inversion (HMMI) to estimate the visual parameters from speech. Zhang et al. \cite{zhang2013new} used a DNN to map speech features into HMM states, which further maps to generated faces.

In recent years, a few DNN-based approaches have also been proposed. Suwajanakorn et al. \cite{suwajanakorn2017synthesizing} designed an LSTM network to generate photo-realistic talking face videos of a target identity directly from speech. Their system requires several hours of face videos of the specific target identity, which greatly limits its application in many practical scenarios. Chung et al. \cite{chung2017you} proposed a convolutional neural network (CNN) system to generate a photo-realistic talking face video from speech and a single face image of the target identity. Compared to \cite{suwajanakorn2017synthesizing}, the reduction from several hours of face videos to a single face image for learning the target identity is a great advance.

While end-to-end speech-to-face-video generation is very useful in many scenarios, the main limitation of this approach is the lack of freedom for further manipulation of the generated face video. For example, within a generated video, one may want to vary the gestures, facial expressions, and lighting conditions, all of which can be relatively independent of the content of the speech. These end-to-end systems cannot accommodate such manipulations unless they can take these factors as additional inputs. However, that would significantly increase the amount and diversity of data required for training the systems.

A modular design that separates the generation of key parameters and the fine details of generated face images is more flexible for such manipulations. Ideally, the key parameters should just respond to the speech content, while the fine details should incorporate all other non-speech-content related factors. Pham et al. \cite{pham2017speech} adopted a modular design: the system first maps speech features to 3D deformable shape and rotation parameters using an LSTM network, and then generates a 3D animated face in real-time from the predicted parameters. In \cite{pham2017end}, they further improved this approach by replacing speech features with raw waveforms as the input and replacing the LSTM network with a convolutional architecture. However, compared to face landmarks used in our proposed approach, these shape and rotation parameters are less intuitive, and the mapping from these parameters to a certain gesture or facial expression is less clear. In addition, the landmarks generated by our system are for a normalized mean face instead of a certain target identity. This also helps remove factors that are not directly related to the voice.

\begin{figure}[t!]
\centering
\includegraphics[width=0.5\linewidth]{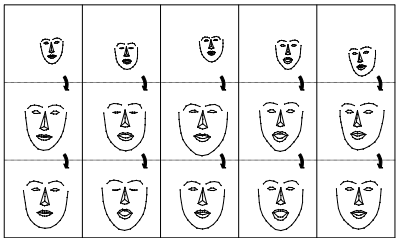}
\caption{Examples of extracted face landmarks from the training talking face videos. Certain landmarks are connected to make the shape of the face easier to recognize. The first row shows unprocessed landmarks of five unique talkers. The second row shows their landmarks after outer-eye-corner alignment. The third row shows their landmarks after alignment and the removal of identity information.}
\label{fig:FlanmarkTransfer}
\end{figure}

\section{Proposed Method} \label{sec:proposed}

	In this section, we describe our method to generate talking face landmarks. First, we extract face landmarks and align them across different speakers and transform their shapes into the mean shape to remove the identity information. We extract the first and second order temporal difference of the log-mel spectrogram and use them as the input to our system. Finally, we train an LSTM network to generate the face landmarks from the speech features.
    
\subsection{Training Data \& Feature Extraction} 

    	We employ the audio-visual GRID dataset \cite{cooke2006audio} to train our system. There are in total 16 female and 18 male native English speakers, each of which has 1000 utterances that are 3 seconds long. The sentences are structured to contain a command, a color, a preposition, a letter, a digit, and an adverb, for example, ``\emph{set blue at C5 please}''.
        
        The videos are provided in two resolutions, low (360x288) and high (720x576). In this work, we use the high-resolution videos. The videos use a frame rate of 25 frames per second (FPS), resulting in 75 frames for each video. The speech audio signal is extracted from the video with a sampling rate of 44.1 kHz.
        
         We extract 68 face landmark points (x and y coordinates) using the DLIB library \cite{dlib09} from each frame for each video in the dataset. Examples are shown in the first row of Figure \ref{fig:FlanmarkTransfer}. We calculate 64 bin log-mel spectra of the speech signal covering the entire frequency range using a 40 ms hanning window without any overlap to match the video frame rate. We then calculate the first- and second-order temporal differences of the log-mel spectra and use them as the input (128-d feature sequence) to our network. We experimented using log-mel spectrogram with and without its first- and second-order derivatives as input to our network. The generated mouth for many speech utterances in these two setups, however, were almost always open even in silent segments, and the lip movements were less prominent than the current system. The first- and second-order temporal differences of the log-mel spectrogram may show less variations on the same syllable uttered by different speakers, and the mismatch problem is less pronounced.
    
\subsection{Face Landmark Alignment} \label{sec:align}

    Since the talking face may appear in different regions with different sizes in different videos, we need to align them to reduce the complexity of training data. To do so, we follow the procedure described in \cite{learnOpenCv} to simply pin the two outer corners of the eyes in the first frame of each video to two fixed locations, (180, 200) and (420, 200) in the image coordinate system, through an 6 DOF affine transformation. We then transform all of the landmarks in all video frames with the same transformation. Note that we do not align each video frame using their own affine transformation separately because we find that the eye-corner-based alignment is sensitive to eye blinks, which often results in zoom in/out effects of the transformed face shape. Also note that our approach assumes that the head does not move significantly within a video, as otherwise, the same affine transformation would not be able to align faces in different frames. The second row of Figure \ref{fig:FlanmarkTransfer} shows several examples of the aligned face landmarks.

\subsection{Removing Identity Information from Landmarks}

After alignment, faces of different speakers are of a similar size and general location; however, their shapes are still different as well as their mouth locations. This identity-related variation may pose challenges to the network for capturing the relation between speech and lip movement, especially when the amount and diversity of training data are small. Therefore, we propose to remove the identity information from the landmarks before training the network.

To do so, we apply the following steps. First, we calculate the mean face shape by averaging all aligned landmark locations across the entire training set. Second, for each face landmark sequence, we calculate the affine transform between the mean shape and the first frame of the sequence. Third, we calculate the difference between the current frame and the first frame and multiply with the scaling coefficients obtained from the second step with the result obtained in the third step. Finally, we add the mean shape to results obtained in fourth step to obtain the face landmark sequence that has no identity. The third row of Figure \ref{fig:FlanmarkTransfer} shows several examples of landmarks with the identity removed.
        
\begin{figure}[t!]
\centering
\includegraphics[width=0.7\linewidth]{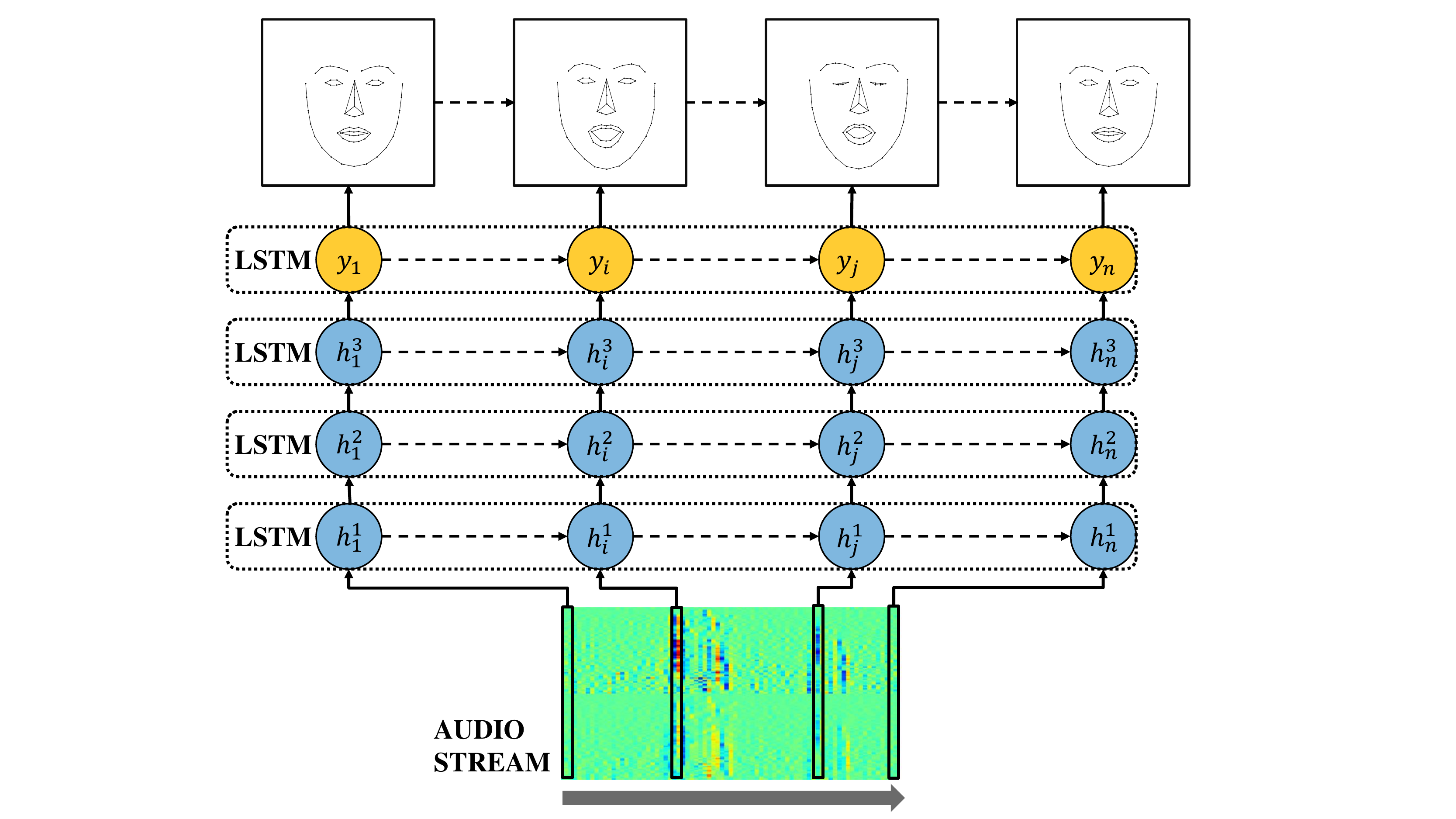}
\caption{The LSTM network architecture for generating landmarks of a talking face from the first and second order temporal differences of the log-mel spectrogram. $h_t^l$ are the hidden layers, where $t$ is the time step and $l$ is the hidden layer index. $y_t$ are the output face landmarks for the time step $t$.}
\label{fig:LSTM}
\end{figure}

\subsection{LSTM Network}    
    Our proposed network, as shown in Figure \ref{fig:LSTM}, uses four long short-term memory (LSTM) \cite{hochreiter1997long} layers with a sigmoid activation function. At each time step, the input to the network is the first and second order temporal differences of the log-mel spectra of the current and the previous N frames. This provides short-term contextual information. The output is the predicted the x and y coordinates of face landmarks of the current frame (if no delay is added) or a previous frame (if a delay is added as described below). The reason for adding delay is because lips often move before the sound is produced. With a little delay, the network is able to ``hear into the future'' and can better prepare for those lip movements. The generated lip movements tend to be smoother. The amount of delay we introduce is between 1 (40 ms) and 5 frames (200 ms). This turns out to be enough for good generation results and is still tolerable in real-time speech communication.
    
    During training, we use dropout between each layer and between recurrent connections, with a rate of 0.2. We use Adam optimizer to train our network. The training sequences are all 75 frames long. We set the batch size to 128 sequences and the learning rate to 0.001. Our network minimizes the following mean squared error (MSE) objective function $J_{MSE}$,   
\begin{equation}\label{cost:1}
  \begin{array}{r c l}
  	J_{MSE} = \frac{1}{N}\sum\limits_{t}^{N} \|\mathbf{s}_t - \widehat{\mathbf{s}}_t\|^2 
  \end{array},
\end{equation}	
\noindent 
  	where $\mathbf{s}$ and $\widehat{\mathbf{s}}$ are the x and y coordinates of ground-truth (GT) and predicted (PD) face landmarks sequences, respectively. $N$ is the number of samples.
   
    	Finally, the predicted landmarks are further processed in order to fix the eye corner points to fixed points as described in Section \ref{sec:align}, which produces more stable talking face landmarks.
        
        Due to causality constraints, the bidirectional LSTM network is not considered in our experiments. We have also experimented with fully connected architecture instead of LSTM. However, the resulting face landmarks often show sudden jumps between frames, which looks unnatural. This is due to not having temporal connections in the architecture.
 
\begin{figure}[t!]
\centering
\includegraphics[width=\linewidth]{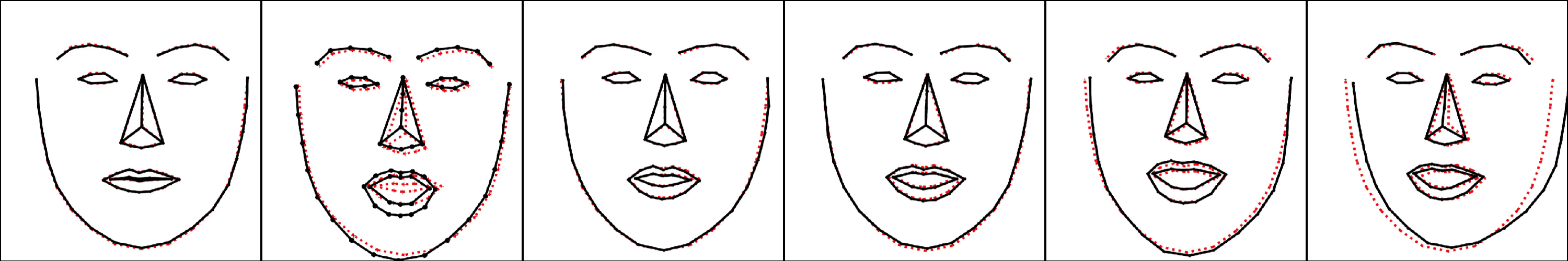}
\caption{Pair-wise comparison between ground-truth landmarks (black solid lines) and generated landmarks (red dotted lines) on unseen talkers and sentences. The second image shows a failure case for ``oh'' sound.}
\label{fig:Ex}
\end{figure}

\section{Experiments}
	 We conduct our objective and subjective evaluations on a totally different audio-visual dataset, the LDC dataset \cite{richie2009audiovisual}. It contains 10 female and 4 male speakers, where each speaker provides 94 samples, totaling to 1316 utterances. The duration of the videos is arbitrary, and the resolution of the samples are 720x480. Since the frame rate of the videos is higher than the Grid dataset used to train our system, we resampled the videos to the same frame rate of 25 FPS. The vocabulary of the LDC dataset is much larger than that of the Grid dataset. There are various words and sentences from TIMIT sentences \cite{garofalo1993darpa}, Northwestern University Auditory Test No. 6 \cite{tillman1966expanded}, and Central Institute for the Deaf (CID) Everyday Sentences \cite{blamey1992factors}. The audio stream is provided at 48 kHz sampling rate, which we down-sampled to 44.1 kHz. Figure \ref{fig:Ex} shows examples of ground-truth and generated face landmarks in the first and second row, respectively. Examples of generated videos are publicly accessible\footnote{\url{http://www.ece.rochester.edu/projects/air/projects/talkingface.html}}.
    
\begin{table}[t!]
\centering
\caption{Objective evaluation results for different system configurations. The models are named according to the amount of delay and contextual information. For example, ``D40-C5'' describes a model trained with 40 ms delay and 5 frames of context. The lower value means better results, where the ideal result is zero. }
\label{tab:results}
\resizebox{0.6\columnwidth}{!}{%
\begin{tabular}{l|c|c|c|}
\cline{2-4}
& \multicolumn{1}{l|}{RMSE} & \multicolumn{1}{l|}{RMSE First Diff} & \multicolumn{1}{l|}{RMSE Second Diff} \\ \hline
\multicolumn{1}{|l|}{D0-C3}           & 0.0954                    & 0.0045                               & 0.0073                                \\ \hline
\multicolumn{1}{|l|}{D0-C5}           & 0.0945                    & 0.0042                               & 0.0071                                \\ \hline
\multicolumn{1}{|l|}{D40-C3}          & 0.0932                    & 0.0039                               & 0.0068                                \\ \hline
\multicolumn{1}{|l|}{\textbf{D40-C5}} & \textbf{0.0921}           & \textbf{0.0032}                      & \textbf{0.0065}                       \\ \hline
\multicolumn{1}{|l|}{D80-C3}          & 0.0946                    & 0.0044                               & 0.0072                                \\ \hline
\multicolumn{1}{|l|}{D80-C5}          & 0.0944                    & 0.0043                               & 0.0069                                \\ \hline
\end{tabular}
}
\end{table}
    
\subsection{Objective Evaluation}

    	We report the root-mean-squared error (RMSE) results between the ground-truth (GT) and predicted (PD) face landmarks according to Equation \ref{cost:1}. The landmarks scale are between 0 and 1, therefore RMSE value of 0.01 approximately equivalent to 1\% error. We also report the RMSE of the first and second order temporal differences of the GT and PD face landmarks to assess the movement. We report the results in Table \ref{tab:results}. These results serve as a way of model selection. The best model according to these results is the model that has 40 ms delay and 5 frames of context information (D40-C5). We selected this model to conduct the subjective evaluations, which are described in the next section.
	
\subsection{Subjective Evaluation}

  We conducted subjective tests to determine if our system can generate realistic face landmarks. 17 naive volunteer evaluators who are graduate students at the University of Rochester participated in the test. The test presented 25 real landmark videos and 25 generated landmark videos in a randomized order to each evaluator and asked the evaluator to label whether each presented video was real or fake. Each video was presented twice in the randomized video sequence. The real landmark videos were created from randomly selected LDC videos. Landmarks were extracted and aligned, and the identity information was removed, according to Section \ref{sec:proposed}. Fake videos were generated from the audio signals of another 25 randomly selected LDC videos. The GT landmarks were noisy; hence we also added Gaussian noise to the PD landmarks to make them look more like the GT landmarks. In addition to a binary decision, the evaluators were asked to report their confidence level of each decision, between 0 and 100 percent.   
    
\begin{figure}[t!]
\centering
\includegraphics[width=0.7\linewidth]{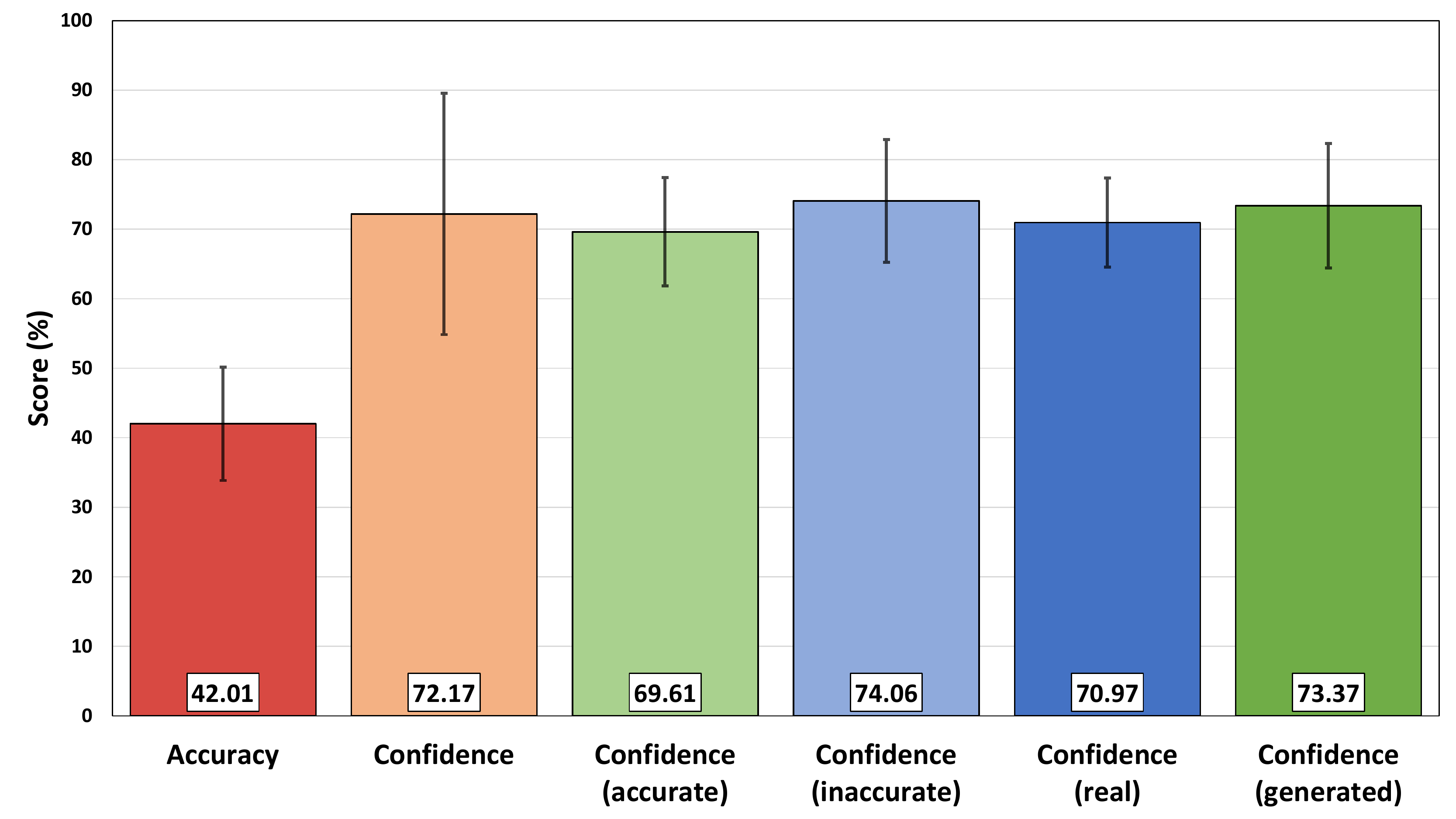}
\caption{Subjective evaluation results. The mean accuracy score and its standard deviation are averaged over all subjects. The mean confidence scores and their standard deviations are averaged over all subjects and videos.}
\label{fig:subjMetrics}
\end{figure}
    
    The mean accuracy score of the evaluators are shown in Figure \ref{fig:subjMetrics}, along with the overall mean confidence score and the mean confidence score for the correctly and incorrectly predicted samples. The results show that the evaluators struggled to distinguish real and generated samples, as the accuracy is 42.01\% which is even below chance (50\%). Another interesting observation of this test is that the mean confidence score for accurately determined samples is lower than that for inaccurately determined samples. This suggests that the evaluators had a higher classification accuracy when they were more cautious. Another outcome is that the mean confidence score on answers for generated samples is more than the confidence score on answers for the ground truth samples. 

\section{Conclusion}

	In this work, we present a method to generate talking face landmarks from speech. We extract face landmarks from the Grid corpus, align them across different speakers, and transform their shapes into the mean shape to remove the identity information. The LSTM network predicts the face landmarks from the first and second order temporal differences of the log-mel spectrogram from any arbitrary voice. The network can produce face landmarks that look natural for the given speech input. The main limitation of this network is that it cannot produce ``oh'' and ``oo'' sounds right. We plan to balance the phonetic content of the dataset to enable the network to produce all phonemes correctly in our future work. We will evaluate the system against noise, and improve it to obtain a noise-resilient system in our future work. We report objective and subjective evaluation results that are promising. We release the code and example videos to the community.

\bibliographystyle{splncs03}
\bibliography{mybib}

\end{document}